\documentclass[lettersize,journal]{IEEEtran}
\usepackage{amsmath,amsfonts}
\usepackage{algorithmic}
\usepackage{algorithm}
\usepackage{array}
\usepackage[caption=false,font=normalsize,labelfont=sf,textfont=sf]{subfig}
\usepackage{textcomp}
\usepackage{stfloats}
\usepackage{url}
\usepackage{verbatim}
\usepackage{graphicx}
\usepackage{cite}

\usepackage{times}
\usepackage{epsfig}
\usepackage{graphicx}
\usepackage{amsmath}
\usepackage{amssymb}
\usepackage{eso-pic}

\usepackage{lineno}
\usepackage{booktabs,makecell, multirow, tabularx}

\usepackage{tikz,xcolor}
\usepackage{hyperref}

\hypersetup{hidelinks,
	colorlinks=true,
	allcolors=black,
	pdfstartview=Fit,
	breaklinks=true}

\begin{document}
\title{Distribution-aware Forgetting Compensation for Exemplar-Free Lifelong Person Re-identification}  
\author{
Shiben Liu \href{https://orcid.org/0000-0001-9376-2562}{\includegraphics[scale=0.08]{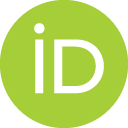}}, 
Huijie Fan
\href{https://orcid.org/0000-0002-8548-861X}{\includegraphics[scale=0.08]{ORCIDiD_icon128x128.png}}, 
Qiang Wang \href{https://orcid.org/0000-0002-2018-1764}{\includegraphics[scale=0.08]{ORCIDiD_icon128x128.png}}, 
Baojie Fan \href{https://orcid.org/0000-0002-1627-9726}{\includegraphics[scale=0.08]{ORCIDiD_icon128x128.png}}
Yandong Tang \href{https://orcid.org/0000-0003-3805-7654}{\includegraphics[scale=0.08]{ORCIDiD_icon128x128.png}}
Liangqiong Qu \href{https://orcid.org/0000-0001-8235-7852}{\includegraphics[scale=0.08]{ORCIDiD_icon128x128.png}}
\thanks{This work is supported by the National Natural Science Foundation of China (62273339, U24A201397), the Key Research and Development Program of Liaoning (2024JH2/102400022), and the LiaoNing Revitalization Talents Program (XLYC2403128). (\emph{Corresponding author: Huijie Fan})}
\thanks{Shiben Liu is with the State Key Laboratory of Robotics, Shenyang Institute of Automation, Chinese Academy of Sciences, Shenyang 110016, China, and with the University of Chinese Academy of Sciences, Beijing 100049, China (e-mail: liushiben@sia.cn).
\par
Huijie Fan, and Yandong Tang are with the State Key Laboratory of Robotics, Shenyang Institute of Automation, Chinese Academy of Sciences, Shenyang, 110016, China (e-mail: fanhuiie@sia.cn; ytang@sia.cn).
\par
Qiang Wang is with the Key Laboratory of Manufacturing Industrial Integrated Automation, Shenyang University, and with the State Key Laboratory of Robotics, Shenyang Institute of Automation, Chinese Academy of Sciences, Shenyang, 110016, China (e-mail: wangqiang@sia.cn). 
\par
Baojie Fan is with the Automation and AI College, Nanjing University of Posts and Telecommunications, Nanjing 210049, China, and also with the State Key Laboratory of Integrated Services Networks, Xi’an 710071, China (e-mail: jobfbj@gmail.com).
\par
Liangqiong Qu is with the Department of Statistics and Actuarial Science and the Institute of Data Science, The University of Hong Kong, Hong Kong, 999077 (e-mail: liangqqu@hku.hk).\\
}%
}

\markboth{}%
{Shell \MakeLowercase{\textit{et al.}}: A Sample Article Using IEEEtran.cls for IEEE Journals}


\maketitle

\begin{abstract}
Lifelong Person Re-identification (LReID) suffers from a key challenge in preserving old knowledge while adapting to new information. The existing solutions include rehearsal-based and rehearsal-free methods to address this challenge. Rehearsal-based approaches rely on knowledge distillation, continuously accumulating forgetting during the distillation process. Rehearsal-free methods insufficiently learn domain-specific distributions, leading to forgetfulness over time. To solve these issues, we propose a novel Distribution-aware Forgetting Compensation (DAFC) model that explores cross-domain shared representation learning, and domain-specific distribution awareness without using old exemplars or knowledge distillation. We propose a Text-driven Prompt Aggregation (TPA) that utilizes text features to enrich prompt components and guide the prompt model to learn fine-grained instance-level representations. This can enhance discriminative identity information and establish the foundation for domain distribution awareness. Then, Distribution-based Awareness and Integration (DAI) is designed to capture each domain-specific distribution by dedicated expert networks and adaptively consolidate them into a shared region in high-dimensional space. In this manner, DAI can consolidate and enhance cross-domain shared representation learning while alleviating catastrophic forgetting. Furthermore, we develop a Knowledge Consolidation Mechanism (KCM) that comprises instance-level discrimination and cross-domain consistency alignment strategies to facilitate model adaptive learning of new knowledge from the current domain and promote knowledge consolidation learning between acquired domain-specific distributions, respectively. Experimental results show that our DAFC outperform state-of-the-art methods by at least 9.8\%/6.6\% and 6.4\%/6.2\% of average mAP/R@1 on two training orders, demonstrating anti-forgetting and generalization capacity. Our code is available at \url{https://github.com/LiuShiBen/DAFC}. 
\end{abstract}
\begin{figure}[t]
\centering
\includegraphics[width=1.0\linewidth, height=0.33\textheight]{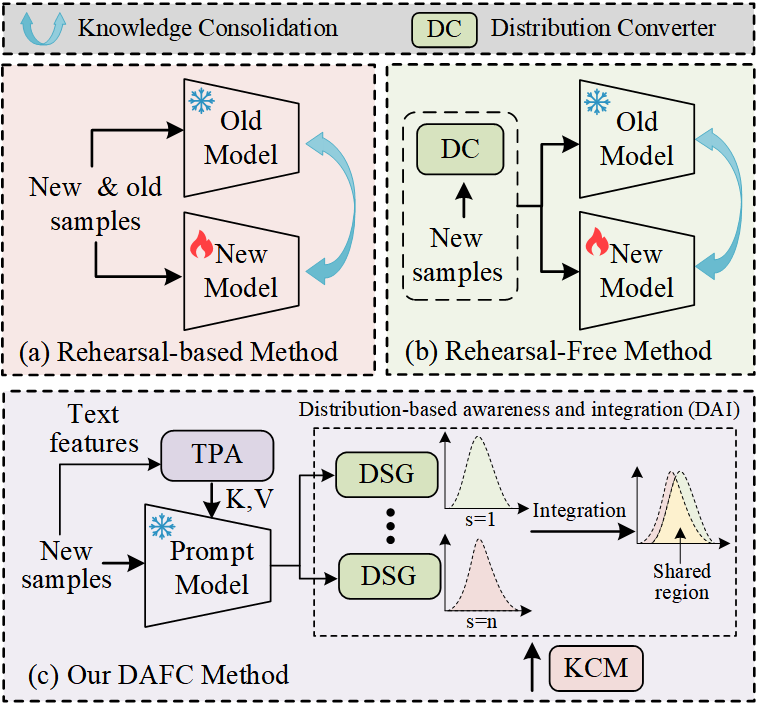}
\caption{The motivation of the proposed DAFC method. (a) Rehearsal-based methods rely on the knowledge distillation strategy, continuously accumulating forgetting during the distillation process and data privacy by storing old exemplars. (b) Rehearsal-free methods utilize the Distribution Converter (DC) to transform the current domain into the style of the last domain, limiting the forgetting ability of the model in the long term. (c) Our DAFC model proposes three innovative components: Text-driven Prompt Aggregation (TPA), Distribution-based Awareness and Integration (DAI), and Knowledge Consolidation Mechanism (KCM) to explore cross-domain shared representation learning and domain-specific distribution awareness, with strong anti-forgetting and generalization capacity without using any old exemplars or knowledge distillation.}
\label{fig:fig1}
\end{figure}
\section{Introduction}
\IEEEPARstart{P}{erson} re-identification (ReID) aims to retrieve individuals across non-overlapping cameras. While existing ReID methods \cite{tao2022dreamt, yang2023, yang2023dual} achieve strong performance within limited time frames or on single datasets, they struggle to adapt to new identities when deployed in scenarios with continuously updated ReID data. As a result, some researchers \cite{chen2024anti, pu2023memorizing, liu2024diverse} have introduced lifelong person re-identification (LReID), extending the domain to an incremental learning setting that evolves over time. A critical challenge in LReID is catastrophic forgetting \cite{li2023cross, cui2025cmoa, pu2021lifelong}, where the model's performance on previously learned domains deteriorates significantly after training on new domains. \\
\indent The LReID methods to address this challenge are primarily rehearsal-based \cite{cui2024learning, xu2024lstkc, yu2023lifelong} and rehearsal-free schemes \cite{li2024exemplar, xu2024distribution, xu2024dask, qian2024auto}. Rehearsal-based methods leverage knowledge distillation strategy to ensure feature consistency between old and new models by using old exemplars and new samples as inputs for training, as illustrated in Fig. \ref{fig:fig1} (a). However, this inherently continues to accumulate forgetting in the distillation process and involves data privacy by storing old exemplars. Rehearsal-free methods \cite{xu2024dask} utilize the Distribution Converter (DC) to transform the current domain into the style of the last domain, as shown in Fig. \ref{fig:fig1} (b). Based on the above LReID methods, there are two main issues to be solved. 1) \textbf{Cross-Domain Shared Representation Learning}. During the incremental learning of new domains, the model struggles to effectively learn representations for all domains. Hence, acquiring fine-grained representations of each instance is cross-domain shared knowledge, which can significantly reduce inter-domain gaps and enhance the model's generalization capability on unseen domains. 2) \textbf{Domain-Specific Distribution Awareness}. The large gaps in distribution between the new and previous domains, primarily attributed to variations in illumination and background, result in the model performing strongly on the new domain but experiencing performance degradation on the previous domain. Therefore, it is crucial to learn domain-specific distributions to enhance the model's capacity against catastrophic forgetting.\\
\indent To solve the above problems, we propose a novel paradigm that explores cross-domain shared representation learning, and domain-specific distribution awareness and integration without relying on knowledge distillation or old exemplars, as illustrated in Fig. \ref{fig:fig1} (c). Existing prompt-based methods \cite{wang2022learning, wang2022s, wang2022dualprompt} in lifelong learning mitigate catastrophic forgetting by learning prompt components for each new domain rather than modifying encoder parameters directly. Inspired by it, we introduce text features with diverse visual attribute information to transform them into prompt components. This enables the prompt model to generate discriminative instance-level representations with cross-domain shared characteristics, reducing inter-domain gaps and laying the foundation for domain-specific distribution perception. Moreover, our method dynamically selects the optimal prompt components, rather than learning per-domain prompt components with fixed shapes. Following Parameter-Efficient Fine-Tuning (PEFT) principles, we design multiple domain-specific generators equipped with dedicated expert networks to effectively capture corresponding domain distribution for preserving old knowledge. Unlike rehearsal-free methods \cite{xu2024distribution, xu2024dask}, our method utilizes previous domain distribution to dynamically adjust current domain distribution for consolidating into a shared region in high-dimensional space, which effectively retains old knowledge and learns cross-domain shared representations.\\
\indent Specifically, we propose a novel Distribution-aware Forgetting Compensation (DAFC) model for LReID that operates without relying on knowledge distillation or old exemplars. Our DAFC consists of three key components: Text-driven Prompt Aggregation (TPA), Distribution-based Awareness and Integration (DAI), and Knowledge Consolidation Mechanism (KCM). The TPA explores the text features to enrich enough prompt components in the domain-shared pool and guide the prompt model to learn fine-grained instance-level representations. It can enhance discriminative identity information and provide rich representations for domain distribution awareness. The domain-shared pool is obtained through text-driven features using a Prompt Generator (PG), which contains diverse visual attribute information as prompt components and dynamically adjusts cross-domain distribution relationships. The PG aims to exploit text features to mine more visual attribute information. We consider that using text features can effectively express visual concepts related to each instance, mining cross-domain shared representations. DAI utilizes previous domain distribution generated by the Domain-Specific Generator (DSG) as old knowledge to consolidate the current domain distribution into a shared region in high-dimensional space. During training in the current domain, the weights from previous domain distributions (learned by DSG) remain frozen for knowledge rehearsal, while only the current domain's DSG is updated to adapt new knowledge. To further enhance learning stability, we develop a KCM that combines instance-level discrimination and cross-domain consistency alignment strategies. This mechanism ensures the adaptive acquisition of new knowledge from the current domain while effectively consolidating knowledge from each domain-specific distribution. Our contributions can be summarized as follows: 
\begin{itemize}
\item[$\bullet$] We propose a novel DAFC model with an exemplar-free and distillation-free framework, which explores domain-specific distribution awareness and integration, and cross-domain shared representations learning for improving anti-forgetting and generalization ability.
\item[$\bullet$] We propose a Text-driven Prompt Aggregation (TPA) that dynamically selects diverse prompt components with diverse visual attribute information and guides the prompt model to learn fine-grained instance-level representations, effectively reducing domain-inter gaps.
\item[$\bullet$] We construct a Distribution-based Awareness and Integration (DAI) to explore domain-specific distribution for knowledge rehearsal and adaptive consolidation into a shared region for alleviating catastrophic forgetting.
\item[$\bullet$] We propose a Knowledge Consolidation Mechanism (KCM) that includes instance-level discrimination and cross-domain consistency alignment strategies to facilitate the adaptive learning of new knowledge and promote inter-domain consistency alignment from each domain-specific distribution, respectively.
\end{itemize}
\section{Related Work}
\subsection{Lifelong Person Re-Identification}
Lifelong Person Re-identification (LReID) has attracted increasing attention due to its significance in handling incremental data streams in real-world scenarios. Simultaneously, the key challenge in the LReID task is how to preserve old knowledge while adapting new knowledge. To solve this challenge, existing LReID approaches \cite{wu2021generalising,pu2021lifelong, ge2022lifelong, sun2022patch, xing2024lifelong, xu2024distribution, xu2024dask} focuse on rehearsal-based and rehearsal-free schemes. Rehearsal-based approaches \cite{pu2021lifelong, yu2023lifelong, huang2023learning, xu2024lstkc} that relys on knowledge distillation strategy, employ the old model as a supervisory signal to achieve feature consistency when using both limited stored samples from previous domains and new samples as training inputs. However, this strategy inherently limits the model's anti-forgetting capability in the distillation process \cite{xu2024dask} and involves data privacy by storing old exemplars. Rehearsal-free methods \cite{xu2024distribution, li2024exemplar, xu2024dask, qian2024auto} retain previous domain distribution as old knowledge prior to preserve knowledge forgetting, limiting the forgetting ability of the model in the long term. Unlike the above methods, we propose a novel exemplar-free model to explore domain-specific distribution awareness and integration, and cross-domain shared representation learning to enhance the model's anti-forgetting and generalization.
\subsection{Mixture of Experts in Lifelong learning}
The Mixture of Experts (MoE) architecture employs a set of specialized expert networks and a gating mechanism to dynamically select the most suitable experts for processing a given input. Several approaches \cite{riquelme2021scaling, mustafa2022multimodal, dai2024deepseekmoe, lin2024moe} integrate MoE frameworks into fine-tuning large language models (LLMs) for downstream tasks, effectively reducing model parameters and enhancing model performance. Some works \cite{zhu2024llama, yu2024boosting, le2024mixture} introduce MoE structure in lifelong learning tasks to prevent catastrophic forgetting issues.  Yu \emph{et al.} \cite{yu2024boosting} proposes a parameter-efficient training framework for vision-language models that integrates Mixture-of-Experts (MoE) to enable experts to simultaneously acquire intra-task knowledge and engage in inter-task collaboration. Le \emph{et al.} \cite{le2024mixture} introduces the attention block of pre-trained models inherently to encode a special mixture of expert architecture. It presents a novel view on prefix tuning, reframing it as the addition of new task-specific experts. These approaches have demonstrated the remarkable performance of MoE architectures in lifelong learning scenarios. Inspired by it, we exploit specialized expert networks to learn domain-specific distribution and consolidate them into a shared region in high-dimensional space, reducing intra-domain gaps during lifelong learning.
\subsection{Prompt Learning}
Prompt-based learning method \cite{wang2022learning, wang2022dualprompt, wang2023isolation,2023progressive,  khan2023introducing, gao2024consistent} effectively addresses catastrophic forgetting by integrating lightweight task-specific instructions (prompts) instead of directly updating the encoder’s parameters. Early approaches, such as L2P \cite{wang2022learning} and Dual-Prompt \cite{wang2022dualprompt}, introduced a shared prompt pool by selecting prompt components through input-driven clustering without requiring explicit task identification. Subsequent advancements like S-Prompts\cite{wang2022s} adapted this framework for domain-incremental learning, where models must handle distribution shifts while retaining fixed class semantics. Further developments include CODA-Prompt \cite{smith2023coda}, which enhances prompt utilization through soft attention mechanisms, enabling end-to-end continual learning. Concurrently, the ConvPrompt \cite{roy2024convolutional} utilized a local prompt creation mechanism that applies convolution over task-invariant global parameters, enabling efficient transfer of local and global concepts across tasks that helps new tasks adapt better. Unlike these methods, we construct text-driven prompt components with diverse visual attribute information, guiding the prompt model to generate fine-grained instance-level representations instead of learning domain-specific prompt components.
\begin{figure*}[!th]
	\centering 
	\includegraphics[width=1\linewidth, height=0.38\textheight]{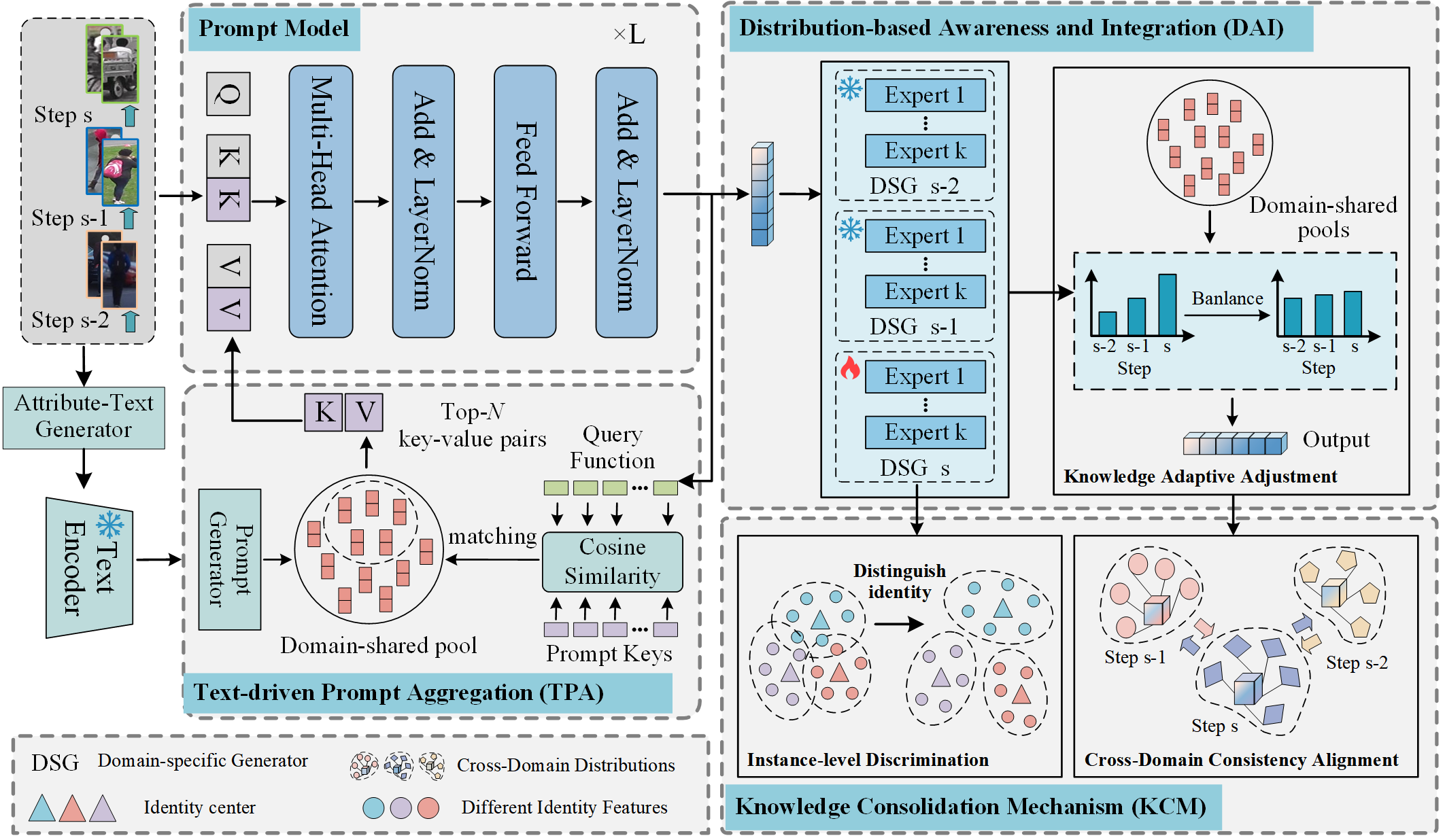}
	\caption{The overall framework of our DAFC.  At the current domain $s$, the attribute-text generator is employed to obtain text-image pairs. Text features are extracted through a pre-trained text encoder and transformed into a domain-shared pool by a Prompt Generator (PG), enhancing the diverse prompt information. Text-driven Prompt Aggregation (TPA) dynamically selects prompt components ($K$, $V$) that are inputted into the prompt model to generate discriminative instance-level representations. Distribution-based Awareness and Integration (DAI) utilizes Domain-Specific Generators (DSG) with dedicated expert networks to capture domain-specific distributions from instance-level representations and consolidate all domain distributions into a shared region in high-dimensional space for reducing inter-domain gaps. Lastly, the Knowledge Consolidation Mechanism (KCM) is designed to facilitate the learning of new knowledge and promote domain consistency alignment.}
	\label{fig:framework}
\end{figure*}
\section{Proposed Method}
\subsection{Preliminary}
In the exemplar-free LReID task, a stream of $S$ domains $D=\{D^s\}_{s=1}^S$ including training domains $\{D_{train}^s\}_{s=1}^S$ and testing domains $\{D_{test}^s\}_{s=1}^S$ are continuously collected to train the LReID model sequentially. $S$ indicates the total number of domains. Unlike traditional LReID methods, we assume that the LReID model cannot access the old exemplars and only utilize the current step's training dataset $D_{train}^s$ in the training process. Given an input of 2D image $x\in\mathbb{R}^{H \times W \times C}$, a prompt model from pre-training vision encoder of CLIP $f_p$, a pre-trained text encoder of CLIP $f_t$ and the classification head $f_c$. To evaluate the anti-forgetting and generalization capabilities of the LReID model, the LReID model trained on the training domains $D$ is directly evaluated on the testing domains $D_{test}$ and a series of unseen domains $U$.
\subsection{Overview of DAFC}
We propose a Distribution-aware Forgetting Compensation (DAFC) model with an exemplar-free and distillation-free framework, as illustrated in Fig. \ref{fig:framework}. Our method consists of an Attribute-Text Generator \cite{liu2024domain}, a Text Encoder, a Prompt model, a Text-driven Prompt Aggregation (TPA), a Distribution-based Awareness and Integration (DAI), and a Knowledge Consolidation Mechanism (KCM). The prompt model aims to learn fine-grained representations for each instance. We introduce the attribute-text generator \cite{liu2024domain} to generate text-image pairs for each instance. Unlike CLIP, text features are generated by a pre-training text encoder without introducing contrastive loss constraints. Text-driven prompt aggregation explores text features to enrich enough prompt components and guide the prompt model to learn fine-grained representations for each instance. Distribution-based awareness and integration explore domain-specific distribution learning and consolidate all domain distribution into a shared region in high-dimensional space.
\subsection{Text-driven Prompt Aggregation}
Learning cross-domain shared representations is crucial in preventing catastrophic forgetting of the LReID task. The existing works \cite{pu2021lifelong, yu2023lifelong} leverage knowledge distillation scheme to ensure domain consistency alignment between old and new models, limiting model anti-forgetting capacity \cite{xu2024dask}. Thus, we propose a Text-driven Prompt Aggregation (TPA) to dynamically select diverse prompt components and guide prompt model to generate fine-grained instance-level representations, including domain shared pool and knowledge matching and maximization.\\
\textbf{Domain shared pool.} Existing approaches\cite{smith2023coda} have shown to perform well by prompting a pre-trained transformer, they rely on a single set of learnable prompt components. Unlike these methods, our method learns prompt-shared components of all domains through text-driven features, instead of prompt-specific components of cross domains by directly learning vectors. Thus, we propose a domain-shared pool generated by Prompt Generator (PG), to provide diverse visual attribute information as prompt components and dynamically adjust cross-domain distribution relationships. The prompt generator exploits text features to mine diverse visual attribute information. We consider that using text features can obtain descriptions of visual attributes of different instances and use these to distinguish similar identity information. Visual attributes are additional semantic knowledge that effectively expresses visual concepts related to each instance, effectively mining cross-domain shared representations. Thus, the Domain-shared pool is proposed to have the following two advantages: 1) It presents diverse visual attribute information as prompt components to provide auto-selected optimal prompt information, guiding prompt model to learn discriminative representations for each instance.. 2) the Domain-shared pool contains rich visual attribute information, which is converted into weight information to adaptive adjust the cross-domain distribution relationship. Specifically, text features $T_{feat}$ are excavated diverse visual attribute information as prompt components in a domain-shared pool through a prompt generator using $M$ linear layers.\\
\begin{equation}
	p_i = \phi_i(T_{feat}) (i=1,2, \cdots, M)
\end{equation}
where $\phi(\cdot)$ refers to linear function. Domain-shared pool include $M$ prompt components with values of $p_i\in\mathbb{R}^{D}$.\\
\textbf{Knowledge Matching and Selection.} To encourage the query mechanism to select diverse prompt components of each instance in the domain-shared pool, we propose knowledge matching and selection strategy. Similar to \cite{smith2023coda}, we replace learnable prompt parameter $P\in\mathbb{R}^{M \times D}$ with a weighted summation over the prompt components of domain-shared pool: 
\begin{equation}
	P = \sum_{i=1}^{M} \gamma(q(x) \scalebox{0.8}{$\bigodot$} A_M, k_M) p_{M} 
\end{equation}
$\gamma(q(x) \scalebox{0.8}{$\bigodot$} A_M, k_M)$ is a weighting vector produced by computing the cosine similarity between a query and keys. Where $A\in\mathbb{R}^{M \times D}$ contains learnable parameters (attention vectors), $k_M$ represents a learnable key $k\in\mathbb{R}^{M \times D}$ to associate keys corresponding to our prompt components from the domain-shared pool. We directly use the whole pre-trained model as a frozen feature extractor to get the query $q(x) = f_p(x)[:, 0, :]$, corresponding to [class].\\
\indent To learn diverse instance-based representations from the domain-shared pool, We dynamically select the top-$N$ prompt parameters from $P$.
\begin{equation}
	\tilde{P} = \text{top-$N$}(P)
\end{equation}
where $\tilde{P}$ is top-$N$ prompt parameter. Unlike CODA \cite{smith2023coda} method, we leverage text-driven features to construct a domian-shared pool as prompt components instead of directly learning the vector. This strategy can enable the model to learn more robust representations for each instance when suffering from domains with significant variation. Additionally, we dynamically select the top-$N$ prompt parameter from $P$ to learn diverse instance-based representations from the domain-shared pool.
\subsection{Prompt Model}
Our approach introduces a prompt model from a pre-trained Visual Encoder of CLIP based on a Vision Transformer (ViT), consisting of stacks of Multi Head Self-Attention (MHSA), layer normalization, and Feed Forward Network (FFN) with residual connections. MHSA block computes self-attention mechanism at $l$th layer is given by:
\begin{equation}
	A(Q_l,K_l,V_l) = \text{softmax}\left(\frac{Q_l K_l^T}{\sqrt{d_k}}\right)V_l \label{SA}
\end{equation}
where $Q_{l}$, $K_{l}$, $V_l$ are query, key, value in self-attention mechanism, respectively.\\
\indent We separate the prompt parameters $\tilde{P}\in\mathbb{R}^{N \times D}$ along dimension $N$, known as key-values pair $\tilde{P}^K_l$, $\tilde{P}^V_l\in\mathbb{R}^{\frac{N}{2}}$. $\frac{N}{2}$ length prefix vectors $\tilde{P}^{(K)}_l$, $\tilde{P}^{(V)}_{l,h}$ are concatenated with the original key $K_l$ and value $V_l$) respectively. Then the self-attention values at each layer $l$ is computed as $A(Q_l, [K_l, \tilde{P}^{K}_l], [V_l, \tilde{P}^{(V)}_l])$ following Eqn.\ref{SA}. In this paper, the prompt model can effectively capture fine-grained representations for each instance, which is conducive to distinguishing pedestrian identities and laying the foundation for learning domain-specific distribution. 
\subsection{Distribution-based Awareness and Integration}
Rehearsal-based methods \cite{pu2021lifelong, huang2023learning, xu2024lstkc} build a memory bank to store old exemplars for alleviating catastrophic forgetting. However, they suffer from data privacy and occupying computing resources. To solve this questions, we propose a Distribution-based Awareness and Integration (DAI) to explore domian-specific distribution learning by Domain-Specific Generator (DSG) and intergrate all domian distribution into a shared reginon in high-dimensional space by Knowledge Adaptive Adjustment (KAA) strategy.\\
\textbf{Domain-specific Generator.} Inspired by the advanced progress of Parameter-Efficient Fine-Tuning (PEFT) methods \cite{yu2024boosting}, we propose a dedicated Domain-Specific Generator (DSG) with dedicated expert networks to learn domain-specific distribution for preserving old knowledge, as shown in Fig. \ref{fig:framework}. Specifically, at current step s, we use $T$ expert networks as a DSG to learn the current domain distribution from instance-level representations $f_p(x)$. The DSG parameters are continuously optimized and updated through training domains $D_{train}^s$ to promote effective aggregation of the current domain's knowledge by the $T$ expert networks. Meanwhile, we freeze all domain-specific generator parameters at the first step s-1, where each domain assigns a dedicated domain-specific generator. The expert network operation is calculated as follows:
\begin{equation}
	G_j = \phi_j(\psi_j(f_p(x))) (j=1, 2, \cdots, T*s)
\end{equation}
where $\phi$ and $\psi$ refer to linear and Multi-layer Perception (MLP) operations. $G_j$ is a representation obtained by each expert network. Each domain assigns $t$ expert networks, so the $s$ domains have a total of $T*s$ expert networks. \\
\noindent \textbf{Knowledge Adaptive Adjustment}. The domain-specific generator effectively consolidates each domain-domain distribution, but the distribution of these domains is independently separated, which limits the model's ability to anti-forgetting. Thus, we propose a Knowledge Adaptive Adjustment (KAA) strategy that utilizes the distribution of each previous domain and current domain to transfer into a shared region. Specifically, we leverage domain-share pool $p=[p_1, p_2, \cdots, p_M]\in\mathbb{R}^{M \times D}$ to map the weight coefficients of the corresponding expert networks, which dynamically adjust the balance of cross-domain knowledge. The weight coefficients are expressed as:\\
\begin{equation}
	\omega = \delta(\phi(\mu(p)))
\end{equation}
where $\mu$($\cdot$) and $\phi$($\cdot$) are mean function and linear function, repesctively.  $\delta$($\cdot$) indicates softmax function. Then, we employ weight coefficients $\omega \in \mathbb{R}^{s*k}$ to integrate representations from all expert networks into a shared representations, effectively reducing inter-domain gaps. To ensure instance-based discrimination and cross-domain consistency alignment, we derive current domain-specific representations $D^{s} \in \mathbb{R}^{D}$ and the shared representation $D^{all} \in \mathbb{R}^{D}$ of all domains, formulated as: 
\begin{equation}
	\begin{aligned}
	&D^{all} = \sum_{j=1}^{k*s} G_j* \omega^j  \\
	&D^{s} = \sum_{j=(k-1)*s+1}^{k*s} G_j* \omega^j
	\end{aligned}
\end{equation}
\indent Unlike PEFT methods \cite{yu2024boosting}, our DAI leverages a domain-specific generator equipped with dedicated expert networks to consolidate domain-specific distributions, rather than fine-tuning parameters layer-by-layer in a pre-trained model. Our approach offers two key advantages: 1) It effectively consolidates each domain's distribution from the instance-level representations, preserving old knowledge without relying on stored old exemplars. 2) All domain-specific distributions are adaptively integrated into a shared region, enhancing cross-domain knowledge learning and reducing inter-domain gaps.
\subsection{Knowledge Consolidation Mechanism}
To strike a balance between preserving prior knowledge and acquiring new knowledge, we propose a Knowledge Consolidation Mechanism (KCM) to enhance model performance in complex scenarios. The KCM comprises  Instance-level Discrimination and Domain Consistency Alignment.\\
\noindent \textbf{Instance-level Discrimination.} At current step $s$, we assign $T$ expert networks to learn current domain distribution and distinguish inter-identity information. Therefore, we introduce cross entropy loss and triplet loss to constrain current domain-specific representations to optimize the parameters of our DAFC model, which can be expressed as:\\
\begin{equation}
	L^{s}_{\mathrm{CE}} = \frac{1}{B}\sum\limits_{i=1}^{B}y_i\log((f_c(D^{s})_i))
\end{equation}
\begin{equation}
	L^{s}_{\mathrm{Tri}} = max(D^s_p-D^s_n+m, 0)
\end{equation}
Where $B$ is the batch size, and $m$ is the margin, $D^s_p$ and $D^s_n$ are the distances from positive samples and negative samples to anchor samples $D^{s}$.\\
\noindent \textbf{Domain Consistency Alignment.} We introduce cross entropy loss and triplet loss to regularize the shared representations for distinguishing identity information Under the prior knowledge of previous domains. This objective is expressed as:\\
\begin{equation}
	L^{all}_{\mathrm{CE}} = \frac{1}{B}\sum\limits_{i=1}^{B}y_i\log((f_c(D^{all})_i))
\end{equation}
\begin{equation}
	L^{all}_{\mathrm{Tri}} = max(D^{all}_p-D^{all}_n+m, 0)
\end{equation}
where $D^s_p$ and $D^s_n$ are the distances from positive samples and negative samples to anchor samples $D^{all}$. \\
\indent At the current domain (step=$s$), we propose a domain consistency alignment (DCA) to achieve cross-domain similarity by cosine distance, which can be adopted as follow:
\begin{equation}
	\begin{aligned}
		&M_{s-(s-1)} = \gamma(D^{s}, D^{s-1})  \\
		&\vdots   \\
		&M_{s-1} = \gamma(D^{s}, D^{1})
	\end{aligned}
\end{equation}
\begin{equation}
	L_{\mathrm{DCA}} = \frac{1}{s-1}\sum\limits_{i=2}^{s} \mu(M_{s-(s-1)})
\end{equation}
where $\gamma(\cdot,\cdot)$ and $\mu$($\cdot$) are cosine distance and mean function. $M_{s-(s-1)}$ indicates .\\
\indent The overall loss function $L$ of our method is formulated as:
\begin{equation}
	L = L^{all}_{\mathrm{CE}} +L^{all}_{\mathrm{Tri}} + L^{s}_{\mathrm{CE}} + L^{s}_{\mathrm{Tri}} + L_{\mathrm{DCA}}
\end{equation}
\section{Experiments}
\par
\begin{table}[t]
	\centering
	\renewcommand{\arraystretch}{1.3}
	\setlength{\tabcolsep}{5pt}
	\caption{Dataset statistics of the LReID benchmark. Since the sampling procedure results in the numbers of train IDs being all 500, the original numbers of IDs are listed for comparison. '-' denotes that the dataset is not used for training} \label{tab:Table1}
	\begin{tabular}{l|l|l|c|c}  
		\hline
		Type &Datasets &Scale &Train IDs &Test IDs \cr
		\hline
		\multirow{5}{*}{Seen}&
		Market-1501\cite{zheng2015scalable}  &Large &500 (751) &750 \cr
		&CUHK-SYSU\cite{xiao2016end}  &Mid &500 (942) &2900 \cr
		&DukeMTMC\cite{ristani2016performance}  &Large &500 (702) &1110 \cr
		&MSMT17$\_$V2\cite{wei2018person} &Large &500 (1041) &3060 \cr
		&CUHK03\cite{li2014deepreid} &Mid &500 (700) &700 \cr
		\hline
		\multirow{6}{*}{Unseen}&
		VIPeR\cite{gray2008viewpoint} &Small &\makecell[c]{$-$} &316  \cr
		&GRID\cite{loy2010time} &Small &\makecell[c]{$-$} &126 \cr
		&CUHK02\cite{li2013locally} &Mid &\makecell[c]{$-$} &239 \cr
		&Occ$\_$Duke\cite{miao2019pose} &Large &\makecell[c]{$-$} &1100 \cr
		&Occ$\_$REID\cite{zhuo2018occluded} &Mid &\makecell[c]{$-$} &200 \cr
		&PRID2011\cite{hirzer2011person} &Small &\makecell[c]{$-$} &649 \cr
		\hline
	\end{tabular}
\end{table}
\par
\subsection{Experiments Setting}
\textbf{Datasets.} To verify the performance of our method in anti-forgetting and generalization, all experiments are conducted on a challenging benchmark consisting of seen domains and unseen domain. Seen domain is mainly used for training and testing, including Market1501\cite{zheng2015scalable}, CUHK-SYSU \cite{xiao2016end}, DukeMTMC \cite{ristani2016performance}, MSMT17$\_$V2 \cite{wei2018person} and CUHK03 \cite{li2014deepreid}. Following the previous work \cite{pu2021lifelong}, two classic training orders are used to imitate varying domain gaps and assess the model performance. Further, unseen domain is employed to test for demonstrating the generalization of the model, consisting of VIPeR \cite{gray2008viewpoint}, GRID \cite{loy2010time},  CUHK02 \cite{li2013locally}, Occ$\_$Duke \cite{miao2019pose}, Occ$\_$REID \cite{zhuo2018occluded}, and PRID2011 \cite{hirzer2011person}. These unseen domain incorporate variations in shape, lighting conditions, and occlusion to effectively demonstrate the generalization capacity of our method. Dataset statistics of the LReID benchmark are shown in Table \ref{tab:Table1}. \\
\textbf{Implementation Details.} Both our prompt-based visual encoder and text encoder with its weights frozen are built upon a pre-trained CLIP model \cite{radford2021learning}, . All input images are resized to 256$\times$128. We use Adam \cite{kinga2015method} for optimization and train each task for 60 epochs with a batch size of 128. The initial learning rate is set to 5$\times$$10^{-6}$ and is reduced by a factor of 0.1 every 20 epochs for each task. \\
\textbf{Evaluation Metrics.} We evaluate model performance in each individual domain using Mean Average Precision (mAP) and Rank-1 (R@1) accuracy. To further assess the model’s lifelong learning capability, we compute the average mAP and R@1 scores across both seen and unseen domains, respectively, serving as a comprehensive evaluation metric.
\begin{table*}[t]
	\centering
	\renewcommand{\arraystretch}{1.3}
	\setlength{\tabcolsep}{6pt}	
	\caption{Performance comparison with state-of-the-art methods on training order-1. The optimal and suboptimal values are highlighted in red and blue, respectively. Training order-1 is Market-1501$\to$CUHK-SYSU$\to$ DukeMTMC$\to$MSMT17$\_$V2$\to$CUHK03.}
	\label{tab:Table2}
	\begin{tabular}{c|cc|cc|cc|cc|cc|cc|cc}
		\hline
		\multirow{2}{*}{Method}&
		\multicolumn{2}{c|}{Market-1501}&\multicolumn{2}{c|}{CUHK-SYSU}&\multicolumn{2}{c|}{DukeMTMC}&\multicolumn{2}{c|}{MSMT17$\_$V2}&
			\multicolumn{2}{c|}{CUHK03}&\multicolumn{2}{c}{\textbf{Seen-Avg}}&\multicolumn{2}{c}{\textbf{Unseen-Avg}}\\
		\cline{2-15}
		&mAP &R@1 &mAP &R@1 &mAP &R@1 &mAP &R@1 &mAP &R@1 &mAP &R@1 &mAP &R@1\cr
		\hline
		SPD\cite{tung2019similarity}&35.6&61.2&61.7&64.0&27.5&47.1&5.2 &15.5 &42.2 &44.3 &34.4 &46.4 &- &-\\
		LwF\cite{li2017learning} &56.3 &77.1 &72.9 &75.1 &29.6 &46.5 &6.0 &16.6 &36.1 &37.5 &40.2 &50.6 &- &-\\
		CRL\cite{zhao2021continual}&58.0 &78.2 &72.5 &75.1 &28.3 &45.2 &6.0 &15.8 &37.4 &39.8 &40.5 &50.8 &- &-\\
		AKA\cite{pu2021lifelong} &58.1 &77.4 &72.5 &74.8 &28.7 &45.2 &6.1 &16.2 &38.7 &40.4 &40.8 &50.8 &42.0 &39.8 \\
		PTKP\cite{ge2022lifelong} &64.4 &82.8 &79.8 &81.9 &45.6 &63.4 &10.4 &25.9 &42.5 &42.9 &48.5 &59.4 &51.2 &49.1 \\
		PatchKD\cite{sun2022patch} &68.5 &85.7 &75.6 &78.6 &33.8 &50.4 &6.5 &17.0 &34.1 &36.8 &43.7 &53.7 &45.1 &43.3\\
		KRKC\cite{yu2023lifelong} &54.0 &77.7 &83.4 &\textcolor{blue}{\textbf{85.4}} &48.9 & 65.5 &14.1 &33.7 &49.9 &50.4 &50.1 &62.5 &\textcolor{blue}{\textbf{52.7}} &\textcolor{blue}{\textbf{50.8}} \\
		ConRFL\cite{huang2023learning} &59.2 &78.3 &82.1 &84.3 &45.6 &61.8 &12.6 &30.4 &51.7 &53.8 &50.2 &61.7 &- &- \\
		LSTKC\cite{xu2024lstkc} &54.7 &76.0 &81.1 &83.4 &49.4 &66.2 &20.0 &43.2 &44.7 &46.5 &50.0 &63.1 &51.3 &48.9\\
		C2R\cite{cui2024learning} &\textcolor{blue}{\textbf{69.0}} &\textcolor{blue}{\textbf{86.8}} &76.7 &79.5 &33.2 &48.6 &6.6 &17.4 &35.6 &36.2 &44.2 &53.7 &- &-\\
		\hline
		DKP\cite{xu2024distribution} &60.3 &80.6 &\textcolor{blue}{\textbf{83.6}} &\textcolor{blue}{\textbf{85.4}} &51.6 &68.4 &19.7 &41.8 &43.6 &44.2 &51.8 &64.1 &49.9 &46.4\\
		DASK \cite{xu2024dask} &61.2 &82.3 &81.9 &83.7 
		&\textcolor{blue}{\textbf{58.5}} &\textcolor{blue}{\textbf{74.6}} &\textcolor{red}{\textbf{29.1}} &\textcolor{red}{\textbf{57.6}} &46.2 &48.1 &\textcolor{blue}{\textbf{55.4}} &\textcolor{blue}{\textbf{69.3}} &- &-\\
		\hline
		Ours &\textcolor{red}{\textbf{81.7}} &\textcolor{red}{\textbf{90.6}} &\textcolor{red}{\textbf{89.7}} &\textcolor{red}{\textbf{90.9}} &\textcolor{red}{\textbf{64.5}} &\textcolor{red}{\textbf{78.9}} &\textcolor{blue}{\textbf{27.1}} &\textcolor{blue}{\textbf{52.9}} &\textcolor{red}{\textbf{64.0}} &\textcolor{red}{\textbf{66.1}} &\textcolor{red}{\textbf{65.6}} &\textcolor{red}{\textbf{75.9}} &\textcolor{red}{\textbf{63.6}} &\textcolor{red}{\textbf{60.5}} \\							
		\hline
	\end{tabular}

\end{table*}
\begin{table*}[!ht]
	\centering
	\renewcommand{\arraystretch}{1.3}
	\setlength{\tabcolsep}{6pt}
	\caption{Performance comparison with state-of-the-art methods on training order-2. The optimal and suboptimal values are highlighted in red and blue, respectively. Training order-2 is DukeMTMC$\to$MSMT17$\_$V2$\to$Market-1501$\to$ CUHK-SYSU$\to$ CUHK03.}
	\label{tab:Table3}
	\begin{tabular}{c|cc|cc|cc|cc|cc|cc|cc}
		\hline
		\multirow{2}{*}{Method}&
		\multicolumn{2}{c|}{DukeMTMC}&\multicolumn{2}{c|}{MSMT17$\_$V2}&\multicolumn{2}{c|}{Market-1501}&\multicolumn{2}{c|}{CUHK-SYSU}& \multicolumn{2}{c|}{CUHK03}&\multicolumn{2}{c}{\textbf{Seen-Avg}}&\multicolumn{2}{c}{\textbf{Unseen-Avg}}\\
		\cline{2-15}
		&mAP &R@1 &mAP &R@1 &mAP &R@1 &mAP &R@1 &mAP &R@1 &mAP &R@1 &mAP &R@1\cr 
		\hline
		SPD\cite{tung2019similarity} &28.5 &48.5 &3.7 &11.5 &32.3 &57.4 &62.1 &65.0 &43.0 &45.2 &33.9 &45.5 &- &- \\
		LwF\cite{li2017learning} &42.7 &61.7 &5.1 &14.3 &34.4 &58.6 &70.0 &73.0 &34.1 &34.1 &37.2 &48.4 &- &- \\
		CRL\cite{zhao2021continual} &43.5 &63.1 &4.8 &13.7 &35.0 &59.7 &70.0 &72.8 &34.5 &36.8 &37.6 &49.2 &- &- \\
		AKA\cite{pu2021lifelong} &42.2 &60.1 &5.4 &15.1 &37.2 &59.8 &71.2 &73.9 &36.9 &37.9 &38.6 &49.4 &41.3 &39.0\\
		PTKP\cite{ge2022lifelong} &54.8 &70.2 &10.3 &23.3 &59.4 &79.6 &80.9 &82.8 &41.6 &42.9 &49.4 &59.8 &50.8 &\textcolor{blue}{\textbf{48.2}}\\
		PatchKD\cite{sun2022patch} &58.3 &74.1 &6.4 &17.4 &43.2 &67.4 &74.5 &76.9 &33.7 &34.8 &43.2 &54.1 &44.8 &43.3\\
		KRKC\cite{yu2023lifelong} &50.6 &65.6 &13.6 &27.4 &56.2 &77.4 &83.5 &85.9 &46.7 &46.6 &50.1 &61.0 &\textcolor{blue}{\textbf{52.1}} &47.7\\
		ConRFL\cite{huang2023learning} &34.4 &51.3 &7.6 &20.1 &61.6 &80.4 &82.8 &85.1 &49.0 &50.1 &47.1 &57.4 &- &-\\
		LSTKC\cite{xu2024lstkc} &49.9 &67.6 &14.6 &34.0 &55.1 &76.7 &82.3 &83.8 &46.3 &48.1 &49.6 &62.1 &51.7 &49.5\\		
		C2R\cite{cui2024learning} &\textcolor{blue}{\textbf{59.7}} &\textcolor{blue}{\textbf{75.0}} &7.3 &19.2 &42.4 &66.5 &76.0 &77.8 &37.8 &39.3 &44.7 &55.6 &- &-\\
		\hline
		DKP\cite{xu2024distribution} &53.4 &70.5 &14.5 &33.3 &60.6 &81.0 &83.0 &84.9 &45.0 &46.1 &51.3 &63.2 &51.3 &47.8 \\
		DASK \cite{xu2024dask} &55.7 &74.4 
		&\textcolor{blue}{\textbf{25.2}} &\textcolor{blue}{\textbf{51.9}} 
		&\textcolor{blue}{\textbf{71.6}} &\textcolor{red}{\textbf{87.7}} &\textcolor{blue}{\textbf{84.8}} &\textcolor{blue}{\textbf{86.2}} &48.4 &49.8 &\textcolor{blue}{\textbf{57.1}} &\textcolor{blue}{\textbf{70.0}} &- &-\\
		\hline
		Ours &\textcolor{red}{\textbf{71.1}} &\textcolor{red}{\textbf{83.2}}
		&\textcolor{red}{\textbf{31.2}} &\textcolor{red}{\textbf{58.0}}
		&\textcolor{red}{\textbf{72.2}} &\textcolor{blue}{\textbf{87.4}} &\textcolor{red}{\textbf{88.2}} &\textcolor{red}{\textbf{89.3}} &\textcolor{red}{\textbf{60.6}} &\textcolor{red}{\textbf{62.9}}
		&\textcolor{red}{\textbf{64.7}} &\textcolor{red}{\textbf{76.2}}
		&\textcolor{red}{\textbf{64.1}} &\textcolor{red}{\textbf{61.4}}\\
		\hline
	\end{tabular}
\end{table*}
\par
\subsection{Comparison with State-of-the-art Methods}
\textbf{Compared Methods.} We implement a detailed comparison between the proposed DAFC and state-of-the-art (SOTA) lifelong person re-identification (LReID) methods to assess their performance in both seen and unseen domains. The comparison approaches are divided into rehearsal-based and Rehearsal-free methods. Rehearsal-based approaches consist of SPD\cite{tung2019similarity}, LwF\cite{li2017learning}, CRL\cite{zhao2021continual}, AKA\cite{pu2021lifelong}, PTKP\cite{ge2022lifelong}, PatchKD\cite{sun2022patch}, KRKC\cite{yu2023lifelong}, ConRFL\cite{huang2023learning}, LSTKC\cite{xu2024lstkc}, C2R\cite{cui2024learning}. Rehearsal-free approaches include DKP\cite{xu2024distribution}, and DASK\cite{xu2024dask}. Experimental results for training order-1 and training order-2 are presented in Table \ref{tab:Table2} and Table \ref{tab:Table3}, respectively. \\
\textbf{Compared with Rehearsal-based Methods.} Rehearsal-based methods adopt a knowledge distillation scheme to ensure feature consistency between old and new models by using stored limiting samples from previous domains and new samples as inputs for training, which often inherently continues to accumulate forgetting in the distillation process. In contrast, our DAFC significantly outperforms rehearsal-based LReID methods, with an seen-avg incremental gain of 15.4\% mAP/12.8\% R@1, and 14.6\% mAP/14.1\% R@1 improvement in Table \ref{tab:Table2} and Table \ref{tab:Table3}, respectively. It can be attributed to utilizing certain domain-specific generators with dedicated expert networks in our DAFC to learn per-domain distributions and aggregate them into a shared region in a high-dimensional region, without relying on knowledge distillation and old exemplars, significantly enhancing its performance from two training orders.\\
\textbf{Compared with Rehearsal-free methods.} Rehearsal-free methods struggle to remain previous domain distribution to preserve old knowledge. They learn insufficient domain distribution due to large gaps across domain, leading to forget old knowledge. Compared with the DKP and DASK methods, our proposed DAFC can significantly outperform them by at least 20.5\%/8.3\%, 6.1\%/5.5\%, 6.0\%/4.3\%, -2.0\%/-4.7\%, and 17.8\%/18\% of mAP/R@1 on seen domains in terms of training order-1, respectively. DASK achieves good performance on MSMT12\_V2 (training order-1) and market-1501 (training order-2), but shows poor performance on the other four domains. We consider that distribution rehearser learning proposed by DASK effectively learn distribution for knowledge rehearsal on the specific datasets, and performs poorly on earlier domains (DukeMTMC of training order-2). In general, our DAFC significantly enhances the representational ability of each instance through text-driven prompt aggregation while effectively alleviating catastrophic forgetting via distribution-based awareness and integration.\\
\begin{figure}[!th]
	\centering 
	\includegraphics[width=1\linewidth, height=0.19\textheight]{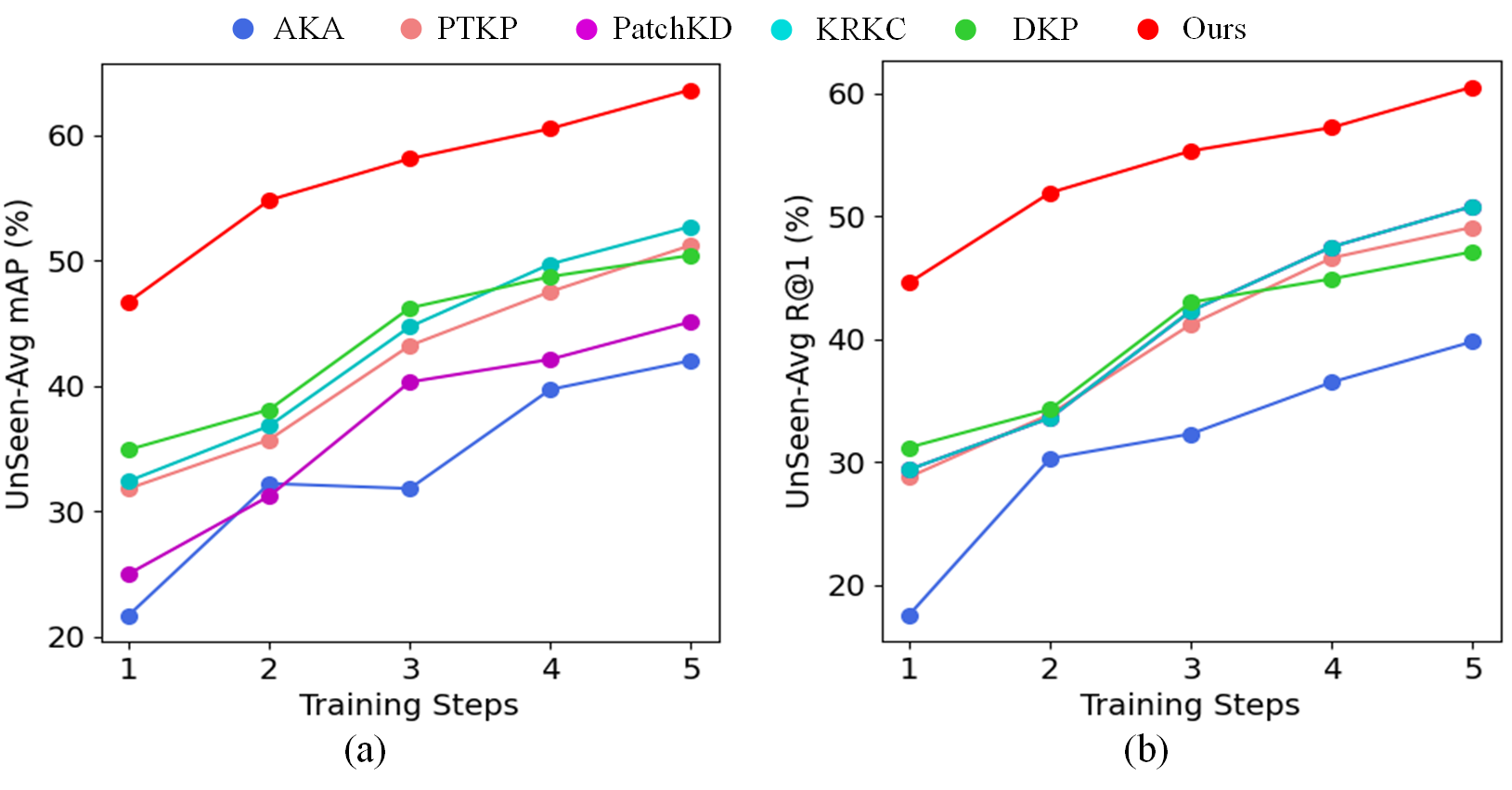}
	\caption{Generalization tendency on unseen domains. (a) mAP metric, (b) R@1 metric.}
	\label{fig:generalization}
\end{figure}\\
\textbf{Generalization of Unseen Domain.} As shown in the Unseen-Avg terms in Table \ref{tab:Table2} and Table \ref{tab:Table3}, the proposed DAFC can significantly outperform them by at least 10.9\%/9.7\%, and 12.0\%/13.2\% of average mAP/R@1 on two training orders. This can be attributed to the domain-based awareness and intergration design that effectively consolidates cross-domain shared knowledge at each training step, guiding the model to learn both intra-domain and inter-domain discriminative representations. To further verify the generalization capability of our DAFC, we provide the average mAP/R@1 results for the unseen domain as the training step increases, as illustrated in Fig. \ref{fig:generalization}. Compared to other methods, some methods experience a slow increase in average mAP/R@1 results as the training steps progress, except for the AKA method where the average mAP result presents a decreasing trend at step 3. Our DAFC model achieves superior performance and shows faster performance growth across the training steps without using a knowledge distillation strategy or storing any old samples. In summary, our DAFC model explores fine-grained representation learning for each instance and learning multiple domain-shared distributions, which significantly enhances generalization ability on the unseen domain.\\
\begin{figure}[!th]
	\centering 
	\includegraphics[width=1\linewidth, height=0.18\textheight]{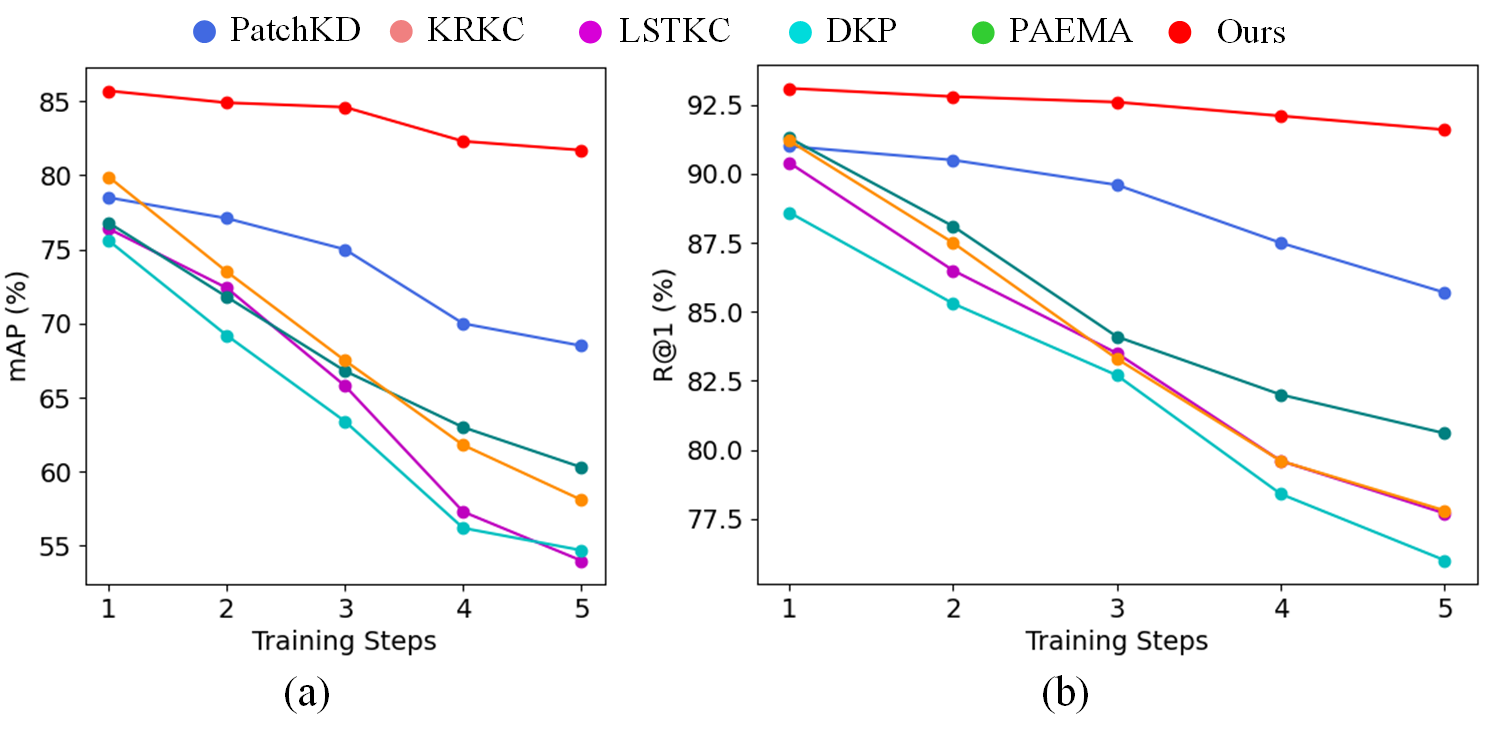}
	\caption{Anti-forgetting curves. (a) mAP metric, (b) R@1 metric. After each training step, we measure the metrics of Market-1501 on training order-1 to demonstrate the model's anti-forgetting performance.}
	\label{fig:anti-forgetting}
\end{figure}\\
\begin{figure*}[t]
	\centering 
	\includegraphics[width=1.0\linewidth, height=0.20\textheight]{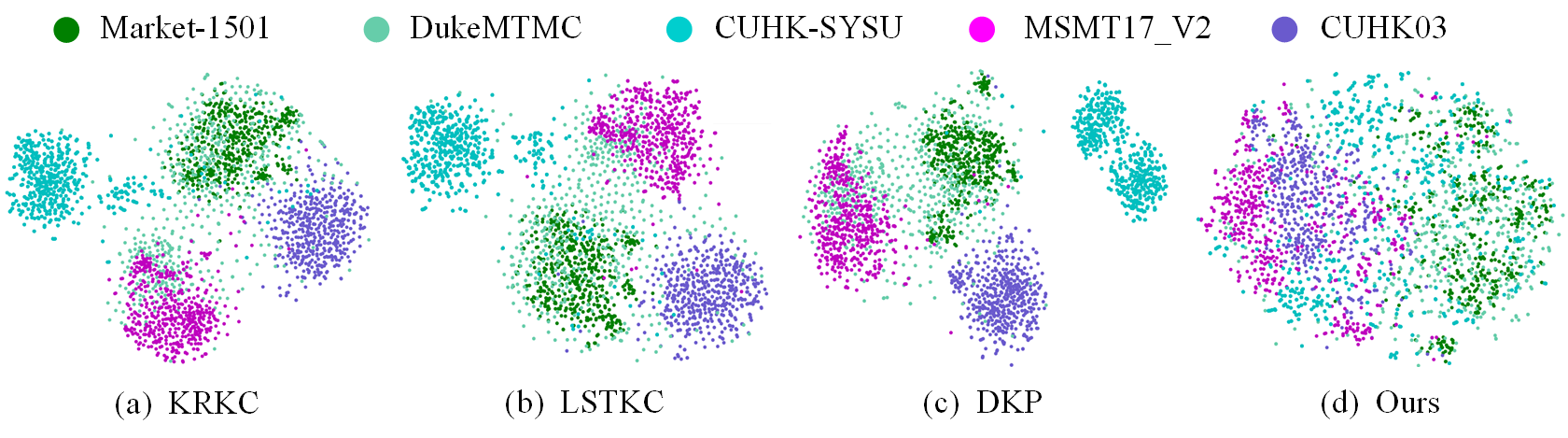}
	\caption{The t-SNE visualization results of the learned representations on five domains.}
	\label{fig:t-sne}
\end{figure*}\\
\textbf{Anti-forgetting Performance Analysis.} To validate the anti-forgetting capability of our method, we measured the metrics on Market-1501 in terms of training order-1 with the number of steps increases, as shown in Fig. \ref{fig:anti-forgetting}. Fig. \ref{fig:anti-forgetting} demonstrates that after training on the large-scale MSMT17 dataset (at step=4), KRKC, LSTKC, and PAEMA methods exhibit significant degradation in both mAP and Rank-1 accuracy, whereas our method demonstrates a more gentle decline. This robustness benefits from our domain-specific generator, which effectively preserves the knowledge distribution of Market-1501 to mitigate forgetting. Compared to other methods, our approach maintains relatively stable mAP and Rank-1 curves as the steps progress, indicating that our method significantly preserves old knowledge through distribution-specific awareness and adaptive integration into a unified distribution space.\\
\textbf{The Influence of Cross-domain Representtaion Learning.} To validate the cross-domain shared representation learning ability of our method, we visualize the representations of KRKC, LSTKC, DKP, and our DAFC across five domains as shown in Fig. \ref{fig:t-sne}. Compared to the KRKC, LSTKC, and DKP methods, the DKP approach is relatively close between four datasets, while the other dataset (DukeMTMC) is far from them, limiting the model's ability to reduce inter-domain gaps. The KRKC and LSTKC methods struggle to accumulate enough cross-domain shared knowledge to pull the datasets closer together. However, they inadequately separate identity information within the domain, which limits the model's ability to prevent forgetting and enhance generalization. Our DAFC not only effectively distinguishes identity information within individual domains but also propagates identity information across all domains. By learning multiple domain-specific distributions in a shared region, it significantly reduces inter-domain gaps while enhancing the model's anti-forgetting capability. This is attributed to the significant learning of refined instance-level representations and the effective consolidation of domain-specific distribution awareness in our DAFC.
\begin{table}[!ht]
	\centering
	\renewcommand{\arraystretch}{1.3}
	\setlength{\tabcolsep}{10pt}
	\caption{Ablation studies of individual components on training order-1.}
	\begin{tabular}{c|c|c|c|c}
		\hline
		\multirow{2}{*}{Components} &\multicolumn{2}{c}{\textbf{Seen-Avg}} &\multicolumn{2}{c}{\textbf{Unseen-Avg}}\\
		\cline{2-5}
		&mAP &R@1 &mAP &R@1 \cr \hline
		w/o PG  &59.6 &67.8 &58.4 &56.1  \\
		w/o DSG &58.3 &65.9 &57.5 &54.8  \\
		w/o KAA &63.5 &72.3 &61.8 &58.6  \\
		W/o CCA &61.4 &70.2 &60.1 &57.9  \\
		Ours    & \textcolor{red}{\textbf{65.6}} &\textcolor{red}{\textbf{75.9}} 
		        &\textcolor{red}{\textbf{63.6}}  &\textcolor{red}{\textbf{60.5}} \\ 
		\hline
	\end{tabular}	
	\label{tab:component}
\end{table}
\begin{table}[!ht]
	\centering
	\renewcommand{\arraystretch}{1.3}
	\setlength{\tabcolsep}{12pt}
	\caption{The number of prompt components in domain-Shared pool for training order-1.}
	\begin{tabular}{c|c|c|c|c}
		\hline
		\multirow{2}{*}{$M$} &\multicolumn{2}{c}{\textbf{Seen-Avg}} &\multicolumn{2}{c}{\textbf{Unseen-Avg}}\\
		\cline{2-5}
		&mAP &R@1 &mAP &R@1 \cr \hline
		15 &64.3 &73.8 &62.4 &59.2  \\
		20 & 65.6 &\textcolor{red}{\textbf{75.9}} 
		&\textcolor{red}{\textbf{63.6}}  &\textcolor{red}{\textbf{60.5}} \\ 
		25 &65.3 &75.2 &62.8 &60.1  \\
		30 &\textcolor{red}{\textbf{65.8}} &75.7 &63.4 &60.3  \\
		\hline
	\end{tabular}	
	\label{tab:prompt}
\end{table}
\begin{table}[!ht]
	\centering
	\renewcommand{\arraystretch}{1.3}
	\setlength{\tabcolsep}{12pt}
	\caption{The number of Top-$N$ key-value pairs in domain-shared pool for training order-1.}
	\begin{tabular}{c|c|c|c|c}
		\hline
		\multirow{2}{*}{$N$} &\multicolumn{2}{c}{\textbf{Seen-Avg}} &\multicolumn{2}{c}{\textbf{Unseen-Avg}}\\
		\cline{2-5}
		&mAP &R@1 &mAP &R@1 \cr \hline
		8 &63.8 &72.4 &62.1 &58.7  \\
		10 & \textcolor{red}{\textbf{65.6}} &\textcolor{red}{\textbf{75.9}} 
		&\textcolor{red}{\textbf{63.6}}  &\textcolor{red}{\textbf{60.5}} \\ 
		12 &64.9 &74.1 &63.1 &60.3  \\
		14 &65.2 &75.5 &63.2 &59.8 \\ 
		\hline
	\end{tabular}	
	\label{tab:top-N}
\end{table}
\begin{table}[!ht]
	\centering
	\renewcommand{\arraystretch}{1.3}
	\setlength{\tabcolsep}{12pt}
	\caption{The number of experts on domain-specific generator for training order-1.}
	\begin{tabular}{c|c|c|c|c}
		\hline
		\multirow{2}{*}{$k$} &\multicolumn{2}{c}{\textbf{Seen-Avg}} &\multicolumn{2}{c}{\textbf{Unseen-Avg}}\\
		\cline{2-5}
		&mAP &R@1 &mAP &R@1 \cr \hline
		3 &62.5 &70.4 &61.7 &58.4  \\
		4 &64.2 &73.8 &62.6 &59.7  \\
		5 & \textcolor{red}{\textbf{65.6}} &75.9 
		&\textcolor{red}{\textbf{63.6}}  &\textcolor{red}{\textbf{60.5}} \\ 
		6 &65.5 &\textcolor{red}{\textbf{76.1}} &62.9 &59.4  \\
		\hline
	\end{tabular}	
	\label{tab:experts}
\end{table}
\subsection{Ablation Studies}
To assess the contributions of individual components in our DAFC, we conduct ablation studies to determine the impact of Prompt Generator (PG), Domain-specific Generator (DSG), Knowledge Adaptive Adjustment (KAA), and Cross-Domain Consistency Alignment (CCA) on model performance. These experiments are performed on unseen and seen domains under training order-1. The components are set as follows: 1) w/o PG: removing the prompt generator, and using a learnable vector in the domain-shared pool; 2) w/o DSG: excluding the domain-specific generator from previous domains, and only the remaining domain-specific generator of the current domain; 3) w/o KAA: without using knowledge adaptive adjustment; 4) w/o CAA: removing cross-domain consistency alignment strategies. This ablation framework isolates the impact of each component for effective validation.\\
\indent As illustrated in Table \ref{tab:component}, The w/o PG results demonstrate that our prompt generator effectively converts text features into multiple prompt components within a domain-shared pool, providing fine-grained representations for each instance while significantly distinguishing identity information. To better preserve knowledge from previous domains without storing any old samples (second row), we design a domain-specific generator that significantly aggregates each domain's distribution in high-dimensional space, thereby enhancing the model's anti-forgetting capability. Compared with w/o KAA, the KAA module adaptively adjusts the distributions of both current and previous domains into a unified region in high-dimensional space, facilitating the learning of cross-domain shared representations. Similarly, compared with w/o CCA, our CAA strategy achieves domain consistency alignment to reduce inter-domain gaps. In summary, each independent component in our DAFC model plays a crucial role. When four components are combined together, our DAFC significantly enhance both the anti-forgetting capacity and generalization ability of the model.\\
\subsection{Analysis of Hyperparameters}
\textbf{The Number of Prompt Components on Domain-Shared Pool.} The Prompt Components are generated by a prompt generator using text-driven features, which encoders diverse visual attribute information. These prompt components effectively provide discriminative key-value pairs to guide the prompt model in learning fine-grained instance-level representations. To analyze the impact of the number of prompt components, we conduct experiments with varying sizes of the domain-shared pool $M$, as shown in Table \ref{tab:prompt}. We observe that increasing $M$ from 15 to 20 leads to incremental improvements in mAP/R@1 performance. However, when $M$ exceeds 20, the results slightly degrade. Thus, our method achieves optimal performance when $M$ is set to 20.\\
\textbf{The Number of Top-$N$ Key-Value Pairs on Domain-Shared Pool.} 
The key-value pairs fine-tune the prompt model to generate discriminative instance-level representations. However, an excessively large number of key-value pairs is unnecessary. We argue that selecting the top-N most relevant key-value pairs is sufficient to provide differentiated representations for each instance. As shown in Table \ref{tab:top-N}, our experiments demonstrate that setting $N$=3 yields the best performance for our method.\\ 
\textbf{The Number of Expert Networks on Domain-Specific Generator.} 
The Domain-Specific Generator employs specialized expert networks to learn domain-specific distributions, enabling the preservation of old knowledge without relying on old exemplars. To evaluate the impact of the number of expert networks ($k$), we conduct experiments as summarized in Table \ref{tab:top-N}. Our results show that as $k$ increases from 3 to 5, the performance of the DAFC model improves consistently. While setting 
$k$=6 achieves the optimal mAP of 76.1 on seen domains, we find that $k$=5 strikes the best balance between model performance and complexity.\\ 
\section{Conclusions}
In this paper, we propose a novel Distribution-aware Forgetting Compensation (DAFC) model for LReID, which consolidates domain-specific distribution and learns cross-domain shared representations for improving anti-forgetting and generalization ability without relying on knowledge distillation or old exemplars. The proposed Text-driven Prompt Aggregation (TPA) leverage text-driven features as prompt components to guide the prompt model for learning fine-grained instance-level representations. Meanwhile, a Distribution-based Awareness and Integration (DAI) is designed to explicitly learns and consolidates domain-specific distributions for mitigating catastrophic forgetting. To further enhance model's stability, we develop a Knowledge Consolidation Mechanism (KCM) facilitates the adaptive learning of new knowledge and improves domain consistency alignment from each domain-specific distribution, respectively. Experimental results demonstrate that our DAFC significantly outperforms state-of-the-art LReID methods under two training orders and exhibits strong generalization in the unseen domain.

\bibliographystyle{IEEEtran}
\bibliography{document.bib}
\vspace{11pt}

\vspace{-33pt}
\begin{IEEEbiography}[{\includegraphics[width=1in,height=1.25in,clip,keepaspectratio]{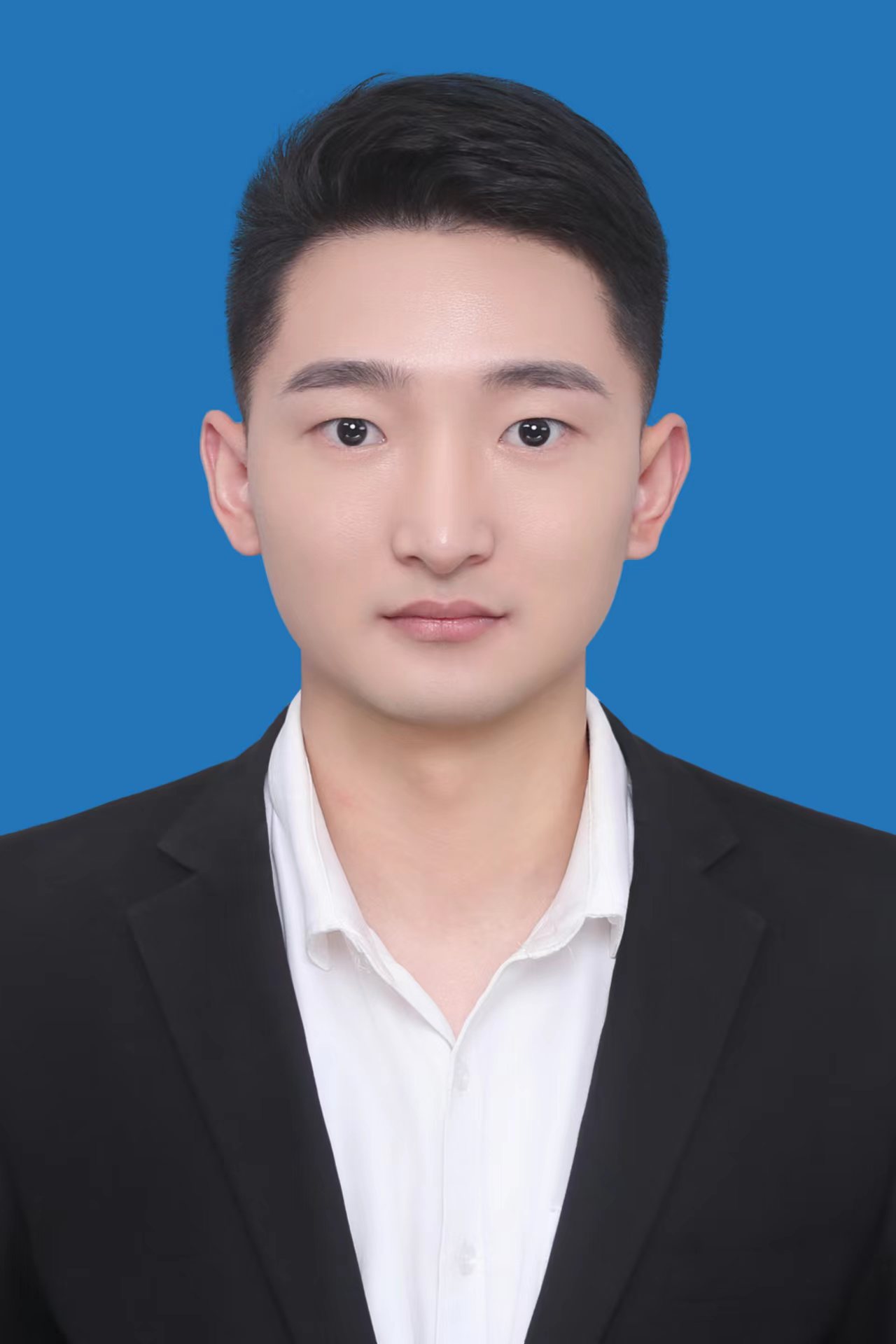}}]{Shiben Liu} received his B.E. and M.S. degrees in Electronic Information Engineering, and Communication and Information Systems from Liaoning University of Engineering and Technology, China, in 2019 and 2022, respectively. He is currently pursuing a Ph.D. degree in State Key Laboratory of Robotics, Shenyang Institute of Automation, University of Chinese Academy of Sciences, Beijing, China. His current research focuses on deep learning, lifelong learning, person re-identification, image restoration and analysis. He has served	as the Reviewer for the international journals such as the TCSVT, TMM, RAL, Neurocomputing, and so on.	
\end{IEEEbiography}
\vspace{11pt}

\vspace{-33pt}
\begin{IEEEbiography}[{\includegraphics[width=1in,height=1.25in,clip,keepaspectratio]{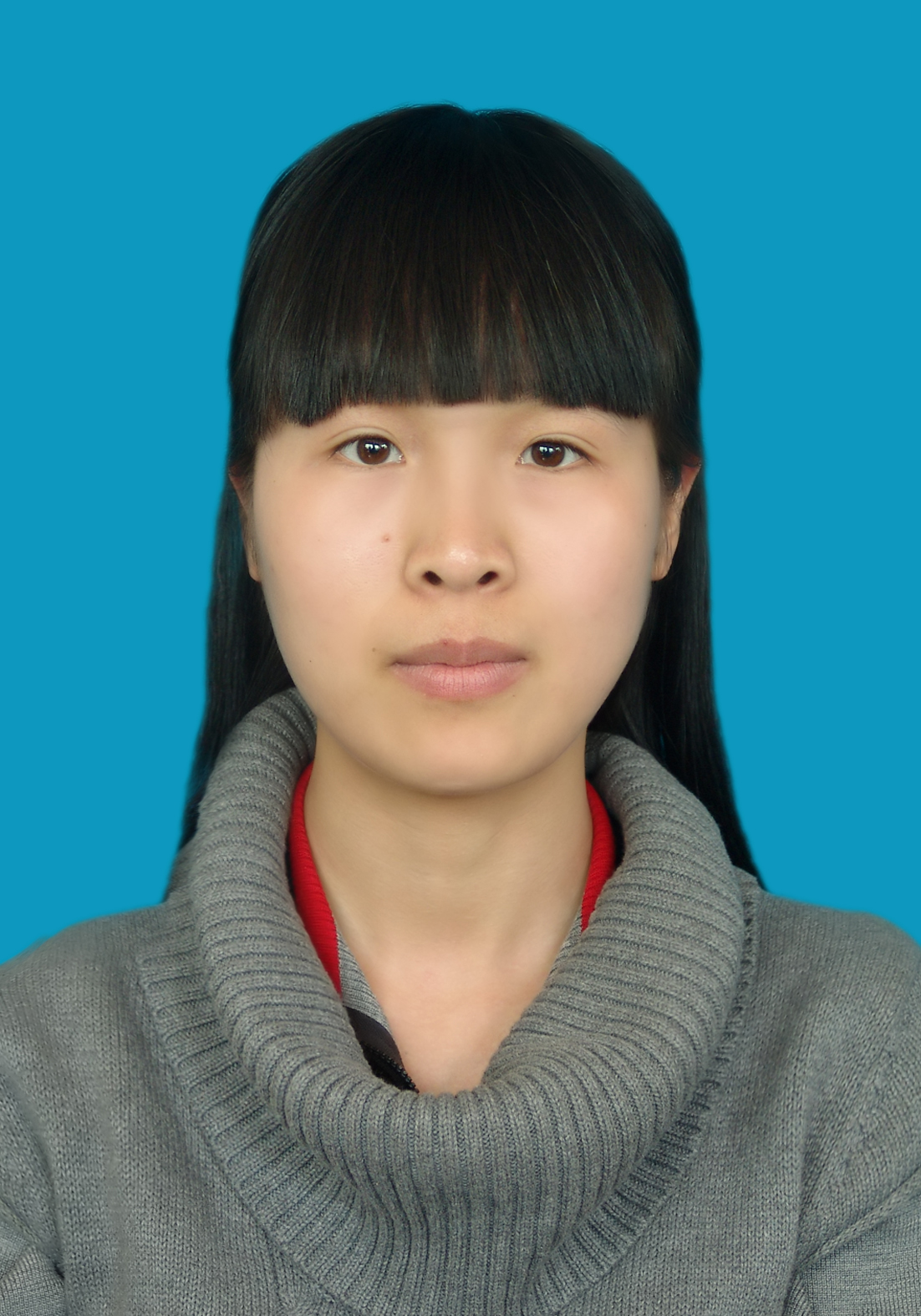}}]{Huijie Fan} (Member, IEEE) received the B.S. degree in automation	from the University of Science and Technology of Science and Technology of China, China, in 2007, and the Ph.D. degree in mode recognition and intelligent systems from the Chinese Academy of Sciences University, Beijing, China, in 2014. She is currently a Research Scientist with the Institute of Shenyang Automation of the Chinese Academy of Sciences. Her research interests include deep learning on image processing and medical image processing and applications.
\end{IEEEbiography}
\vspace{11pt}

\vspace{-33pt}
\begin{IEEEbiography}[{\includegraphics[width=1in,height=1.25in,clip,keepaspectratio]{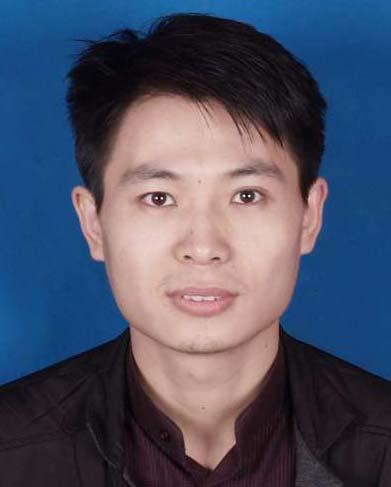}}]{Qiang Wang} received the B.E. and M.S. degrees in school of Computer Science and Technology from Shandong Jianzhu University and Tianjin Normal University, P.R.China,in 2004 and 2008, respectively, and the Ph.D degree in State Key Laboratory of Robotics, Shenyang Institute of Automation, University of Chinese Academy of Sciences, Beijing, China in 2020. He is an Associate Professor in the Key Laboratory of Manufacturing Industrial Integrated in Shenyang University. He has some top-tier journal papers accepted at TIP, TMM, TCSVT, IoT-J and Pattern Recognition et al. His current research focuses on deep learning, multi-task learning, image restoration and analysis.
\end{IEEEbiography}
\vspace{11pt}

\vspace{-33pt}
\begin{IEEEbiography}[{\includegraphics[width=1in,height=1.25in,clip,keepaspectratio]{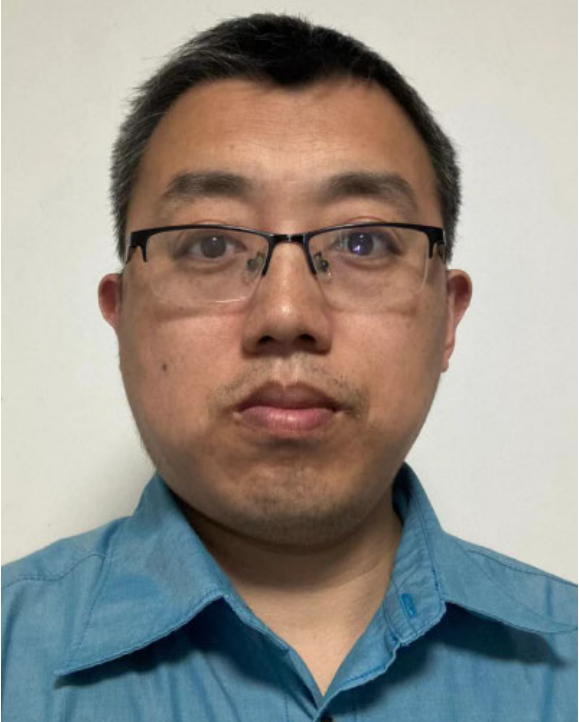}}]{Baojie Fan} received the Ph.D. degree in pat- tern recognition and intelligent system from the State Key Laboratory of Robotics,Shenyang Insti- tute Automation,Chinese Academy of Sciences. He is currently a Professor with the Department of Automation, NJUPT. His major research interests include robot vision systems, object detection, and tracking.
\end{IEEEbiography}
\vspace{11pt}

\vspace{-33pt}
\begin{IEEEbiography}[{\includegraphics[width=1in,height=1.25in,clip,keepaspectratio]{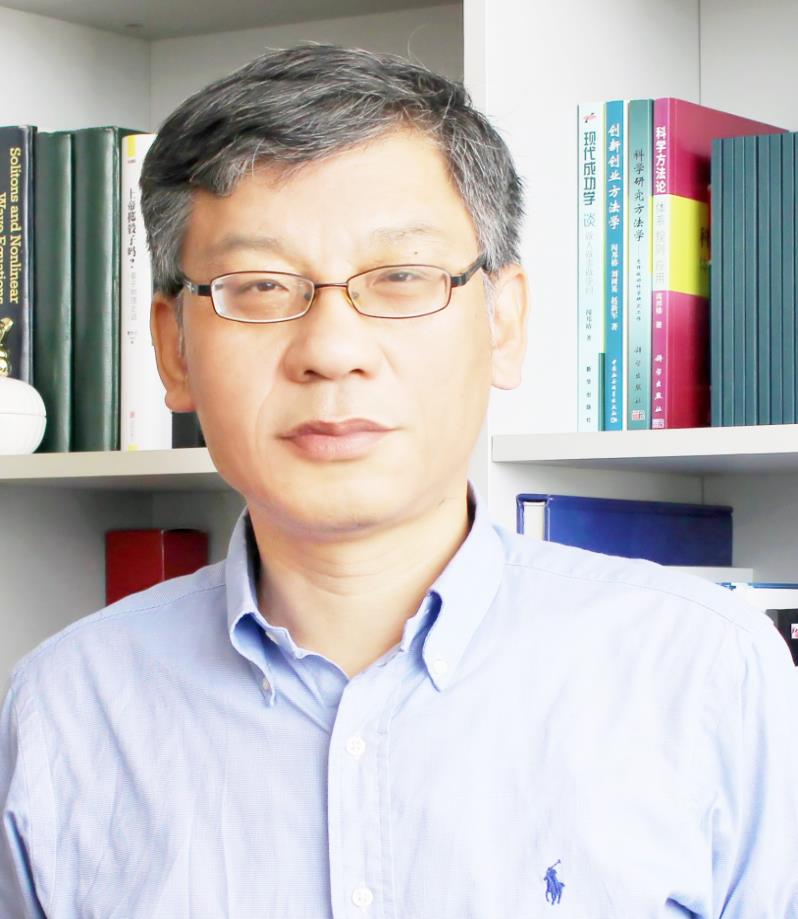}}]{Yandong Tang} received the B.S. and M.S. degree	in the calculation of mathematics from Shandong University, China, in 1984 and 1987. From 1987 to 1996, he worked at the Institute of Computing Technology, Shenyang, Chinese Academy of Sciences. From 1996 to 1998, he was engaged in research and development at Stuttgart University	and Potsdam University in Germany. He received a Ph.D. degree in Engineering Mathematics from the	Research Center (ZETEM) of Bremen University, Germany, in 2002. From 2002 to 2004, he worked at the Institute of Industrial Technology and Work Science (BIBA) at Bremen	University of Germany. He is currently a Research Scientist with the Institute of Shenyang Automation of the Chinese Academy of Sciences. His research interests include image processing, mode recognition and robot vision.
\end{IEEEbiography}
\vspace{11pt}

\vspace{-33pt}
\begin{IEEEbiography}[{\includegraphics[width=1in,height=1.25in,clip,keepaspectratio]{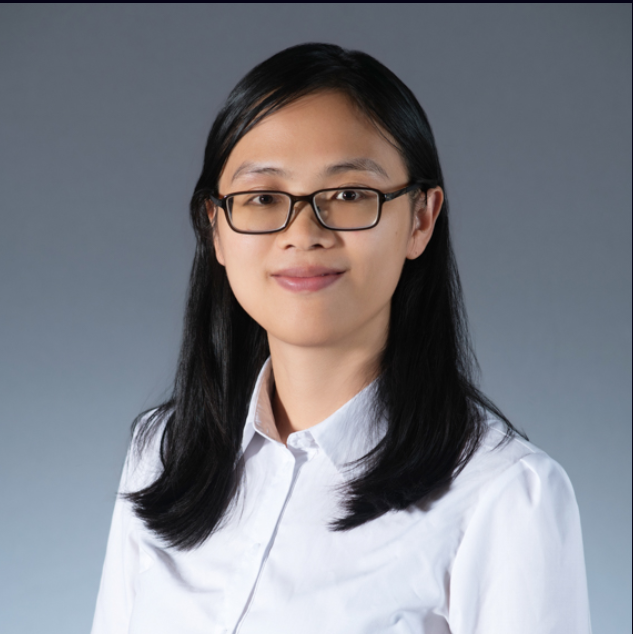}}]{Liangqiong Qu} received the Ph.D.degree in pattern recognition and intelligent system from the Uni- versity of ChineseAcademy of Sciences, Beijing, China, in 2018, and the Ph.D.degree in computer science from the City University of Hong Kong, Hong Kong,in 2018.She was a Post-Doctoral Re- search Fellow with The University of North Carolina at Chapel Hill, Chapel Hill, NC, USA,from 2018 to 2019, and a Post-Doctoral Research Fellow with Stanford University, Stanford, CA, USA.She is cur- rently an Assistant Professor with the Department of Statistics and Actuarial Science, The University of Hong Kong.Her research interests include computer vision, machine learning, and deep learning, with a focus on applications to health care.
\end{IEEEbiography}

\vfill
\end{document}